\documentclass[10pt,twocolumn,letterpaper]{article}

\usepackage{cvpr}
\usepackage{times}
\usepackage{epsfig}
\usepackage{graphicx}
\usepackage{amsmath}
\usepackage{amssymb}
\usepackage{cite}

\usepackage{float}
\usepackage{multirow}
\usepackage{array}
\usepackage{booktabs}
\usepackage{flushend}
\usepackage{marvosym}
\cvprfinalcopy 

\def\cvprPaperID{2852} 

\ifcvprfinal\fi
\begin{document}

\title{Disentangling Features in $3$D Face Shapes \\for Joint Face Reconstruction and Recognition\thanks{This work is supported by the National Key Research and Development Program of China (2017YFB0802300) and the National Natural Science Foundation of China (61773270, 61703077).}}


\author{Feng Liu$^1$, Ronghang Zhu$^1$, Dan Zeng$^1$, Qijun Zhao$^{1,}$\thanks{Corresponding author. Email: qjzhao@scu.edu.cn.}~,~~and Xiaoming Liu$^2$\\
$^1$College of Computer Science, Sichuan University\\
	$^2$Department of Computer Science and Engineering, Michigan State University\\	
{\tt\small }
\and
\\
}

\maketitle

\begin{abstract}
This paper proposes an encoder-decoder network to disentangle shape features during $3$D face reconstruction from single $2$D images, such that the tasks of reconstructing accurate $3$D face shapes and learning discriminative shape features for face recognition can be accomplished simultaneously. Unlike existing $3$D face reconstruction methods, our proposed method directly regresses dense $3$D face shapes from single $2$D images, and tackles identity and residual (i.e., non-identity) components in $3$D face shapes explicitly and separately based on a composite $3$D face shape model with latent representations. We devise a training process for the proposed network with a joint loss measuring both face identification error and $3$D face shape reconstruction error. To construct training data we develop a method for fitting $3$D morphable model ($3$DMM) to multiple $2$D images of a subject. Comprehensive experiments have been done on MICC, BU$3$DFE, LFW and YTF databases. The results show that our method expands the capacity of $3$DMM for capturing discriminative shape features and facial detail, and thus outperforms existing methods both in $3$D face reconstruction accuracy and in face recognition accuracy.
\end{abstract}

\section{Introduction}

\begin{figure}[t]
\centering
\includegraphics[width=0.98\linewidth]{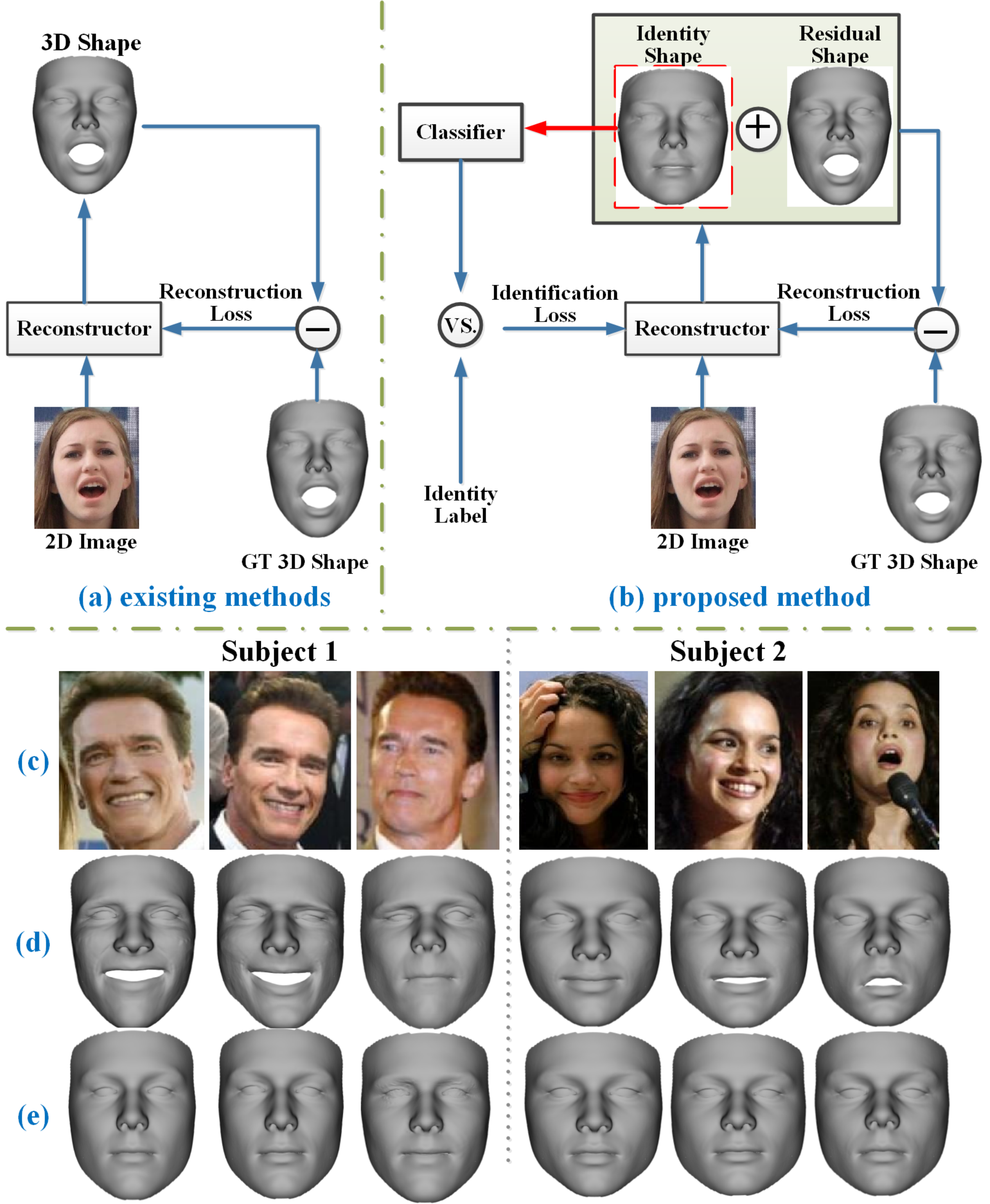}
\caption{Comparison between the learning process of (a) existing methods and (b) our proposed method. GT denotes Ground Truth. (d) and (e) are $3$D face shapes and disentangled identity shapes reconstructed by our method for the images in (c) from LFW \cite{huang2007labeled}.}
\label{fig:comparison}
\end{figure}

$3$D face shapes reconstructed from $2$D images have been proven to benefit many tasks, e.g., face alignment or facial landmark localization~\cite{zhu2016CVPR, jourabloo2017ijcv}, face animation~\cite{cao2016real, HanGY17}, and face recognition~\cite{blanz2003face, han20123d}. Many prior work have been devoted to reconstructing $3$D face shapes from a single $2$D image, including shape from shading (SFS)-based methods~\cite{horn1989shape, kemelmacher20113d}, $3$D morphable model ($3$DMM) fitting-based methods~\cite{blanz1999morphable, blanz2003face}, and recently proposed regression-based methods~\cite{liu2015cascaded, liu2016joint}. 
These methods mostly aim to recover $3$D face shapes that are loyal to the input $2$D images or retain as much facial detail as possible (see Fig.~\ref{fig:comparison}). 
Few of them explicitly consider the identity-sensitive and identity-irrelevant features in the reconstructed $3$D faces. 
Consequently, very few studies have been reported about recognizing faces using the reconstructed $3$D face either by itself or by fusing with legacy $2$D face recognition~\cite{blanz2003face, tran2016regressing}.

Using real $3$D face shapes acquired by $3$D face scanners for face recognition, on the other hand, has been extensively studied, and promising recognition accuracy has been achieved \cite{bowyer2006survey, Emambakhsh2016Nasal}. Apple recently claims to use $3$D face matching in its iPhone X for cellphone unlock~\cite{apperID}. All of these prove the discriminative power of $3$D face shapes. Such a big performance gap between the reconstructed $3$D face shapes and the real $3$D face shapes, in our opinion, demonstrates that existing $3$D face reconstruction methods seriously undervalue the identity features in $3$D face shapes. Taking the widely used $3$DMM fitting based methods as example, their reconstructed $3$D faces are constrained in the limited shape space spanned by the pre-determined bases of $3$DMM, and thus perform poorly in capturing the features unique to different individuals \cite{Yin2017Towards}.

Inspired by the latest development in disentangling feature learning for $2$D face recognition \cite{tran2017disentangled, peng2017recons}, we propose to disentangle the identity and non-identity components of $3$D face shapes, and more importantly, fulfill \emph{reconstructing accurate $3$D face shapes} loyal to input $2$D images and \emph{learning discriminative shape features} effective for face recognition in a \emph{joint} manner. These two tasks, at the first glance, seem to contradict each other. On one hand, face recognition prefers identity-sensitive features, but not every detail on faces; on the other hand, $3$D reconstruction attempts to recover as much facial detail as possible, regardless whether the detail benefits or distracts facial identity recognition. In this paper, however, we will show that by exploiting the `contradictory' objectives of recognition and reconstruction, we are able to \emph{disentangle identity-sensitive features from identity-irrelevant features in $3$D face shapes}, and thus simultaneously robustly recognize faces with identity-sensitive features and accurately reconstruct $3$D face shapes with both features (see Fig. \ref{fig:comparison}).

Specifically, we represent $3$D face shapes with a composite model, in which identity and residual (i.e., non-identity) shape components are represented with separate latent variables. Based on the composite model, we propose a joint learning pipeline that is implemented as an encoder-decoder network to disentangle shape features during reconstructing $3$D face shapes. The encoder network converts the input $2$D face image to identity and residual latent representations, from which the decoder network recovers its $3$D face shape. The learning process is supervised by both reconstruction loss and identification loss, and based on a set of $2$D face images with labelled identity information and corresponding $3$D face shapes that are obtained by an adapted multi-image $3$DMM fitting method. Comprehensive evaluation experiments prove the superiority of the proposed method over existing baseline methods in both $3$D face reconstruction accuracy and face recognition accuracy. Our main contributions are summarized below.

(i) We propose a method which for the first time explicitly optimizes face recognition and $3$D face reconstruction simultaneously. The method achieves state-of-the-art $3$D face reconstruction accuracy via joint discriminative feature learning and $3$D face reconstruction.

(ii) We devise an effective training process for the proposed network that can disentangle identity and non-identity features in reconstructed $3$D face shapes. The network, while being pre-trained by $3$DMM-generated data, can surmount the limited $3$D shape space determined by the $3$DMM bases, in the sense that it better captures identity-sensitive and identity-irrelevant features in $3$D face shapes.

(iii) We leverage the effectiveness of disentangled identity features in reconstructed $3$D face shapes for improving face recognition accuracy, as being demonstrated by our experimental results. This further expands the application scope of $3$D face reconstruction.



\section{Related Work}
\begin{figure*}[t]
\begin{center}
\includegraphics[width=0.85\linewidth]{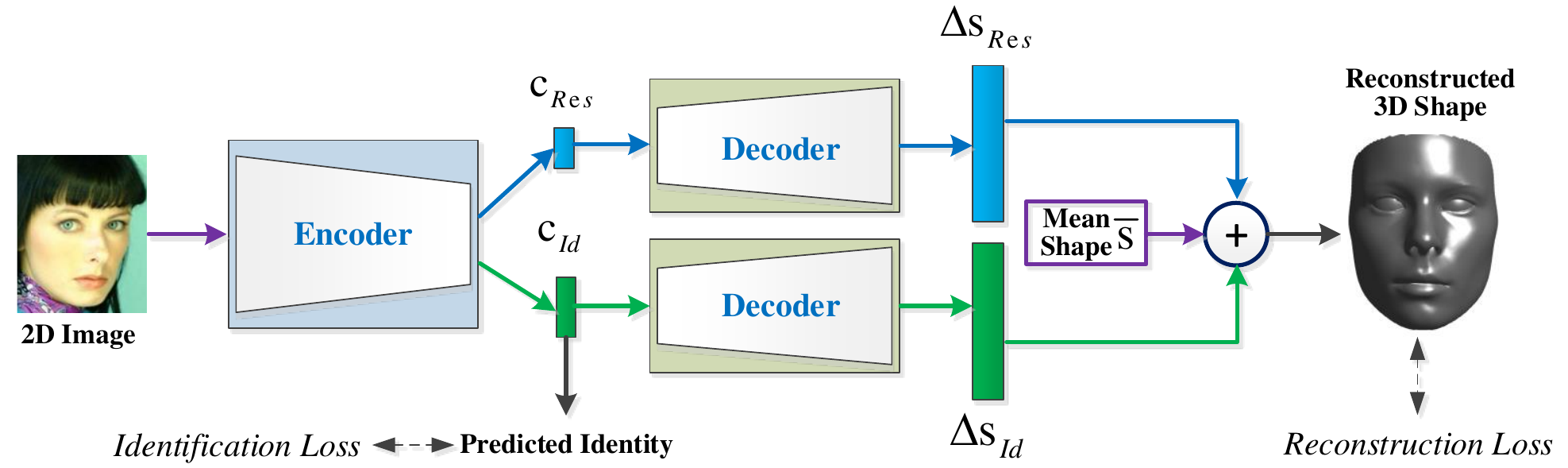}
\end{center}
   \caption{Overview of the proposed encoder-decoder based joint learning pipeline for face recognition and $3$D shape reconstruction.}
\label{fig:framework}
\end{figure*}

In this section, we review existing work that is closely related to our work from two aspects: $3$D face reconstruction for recognition and Convolutional Neural Network (CNN) based $3$D face reconstruction.

\textbf{$3$D Face Reconstruction for Recognition.} $3$D face reconstruction was first introduced for recognition by Blanz and Vetter~\cite{blanz2003face}. They reconstructed $3$D faces by fitting $3$DMM to $2$D face images, and used the obtained $3$DMM parameters as features for face recognition. Their employed $3$DMM fitting method is essentially an image-based analysis-by-synthesis approach, which does not consider the features unique to different individuals. This method was recently improved by Tran et al.~\cite{tran2016regressing} via pooling the $3$DMM parameters of the images of the same subject and using a CNN to regress the pooled parameters. They experimentally proved the improved discriminative power of their obtained $3$DMM parameters.

Instead of using $3$DMM parameters for recognition, Liu et al.~\cite{liu2016joint} proposed to recover pose and expression normalized $3$D face shapes directly from $2$D face landmarks via cascaded regressors and match the reconstructed $3$D face shapes via the iterative closest point algorithm for face recognition. Other researchers~\cite{yi2013towards, taigman2014deepface} utilized the reconstructed $3$D face shapes for face alignment to assist extracting pose-robust features.

To summarize, \emph{existing methods, when reconstructing $3$D face shapes, do not explicitly consider recognition performance}. In~\cite{liu2016joint} and~\cite{tran2016regressing}, even though the identity of $3$D face shapes in the training data is stressed, respectively, by pooling $3$DMM parameters and by normalizing pose and expression, their methods of learning mapping from $2$D images to $3$D face shapes are \emph{unsupervised} in the sense of utilizing identity labels of the training data (see Fig.~\ref{fig:comparison}).

\textbf{CNN-based $3$D Face Reconstruction.} Existing CNN-based $3$D face reconstruction methods can be divided into two categories according to the way of representing $3$D faces. Methods in the first category use $3$DMM parameters~\cite{zhu2016CVPR, Richardson2016Learning, tran2016regressing, Dou2017End, Sela2017Unrestricted, tewari2017mofa}, while methods in the second category use $3$D volumetric representations. Jourabloo and Liu~\cite{jourabloo2015pose,Amin_ICCV17, jourabloo2017ijcv} first employed CNN to regress $3$DMM parameters from $2$D images for the purpose of large-pose face alignment. In~\cite{zhu2016CVPR}, a cascaded CNN pipeline was proposed to exploit the intermediate reconstructed $3$D face shapes for better face alignment. Recently, Richardson et al.~\cite{Richardson2016Learning} used two CNNs to reconstruct detailed $3$D faces in a coarse-to-fine approach. Although they showed visually more plausible $3$D shapes, it is not clear how beneficial the reconstructed  $3$D facial details are to face recognition.

Jackson et al.~\cite{jackson2017vrn} proposed to represent $3$D face shapes by $3$D volumetric coordinates, and train a CNN to directly regress the coordinates from the input $2$D face image. Considering the high dimensionality of original $3$D face point clouds, as a compromise, they employed $3$D volumetric representations. In consequence, the $3$D face shapes generated by their method are of low resolution, which are apparently not favorable for face recognition.


\section{Proposed Method}

In this section, we first introduce a composite $3$D face shape model with latent representations, based on which our method is devised. We then present the proposed encoder-decoder based joint learning pipeline. We finally give the implementation detail of our proposed method, including network structure, training data, and training process.

\subsection{A Composite $3$D Face Shape Model}

In this paper, $3$D face shapes are densely aligned, and each $3$D face shape is represented by the concatenation of its vertex coordinates as
\begin{equation}
\bold{s} = [x_{1}, y_{1}, z_{1}, x_{2}, y_{2}, z_{2}, \cdots, x_{n}, y_{n}, z_{n}]^{T},
\end{equation}
where $n$ is the number of vertices in the point cloud of the $3$D face, and `$T$' means transpose. Based on the assumption that $3$D face shapes are composed by identity-sensitive and identity-irrelevant parts, we re-write the $3$D face shape $\bold{s}$ of a subject as
\begin{equation}
\label{eq:shape_compose}
\bold{s} = \bar{\bold{s}} + \Delta \bold{s}_{Id} + \Delta \bold{s}_{Res},
\end{equation}
where $\bar{\bold{s}}$ is the mean $3$D face shape (computed across all training samples with neutral expression), $\Delta \bold{s}_{Id}$ is the identity-sensitive difference between $\bold{s}$ and $\bar{\bold{s}}$, and $\Delta \bold{s}_{Res}$ denotes the residual difference. A variety of sources could lead to the residual difference, for example, expression-induced deformations and temporary detail.

\begin{figure}[t]
\begin{center}
\includegraphics[trim={1mm 0 1mm 0},clip,width=\linewidth]{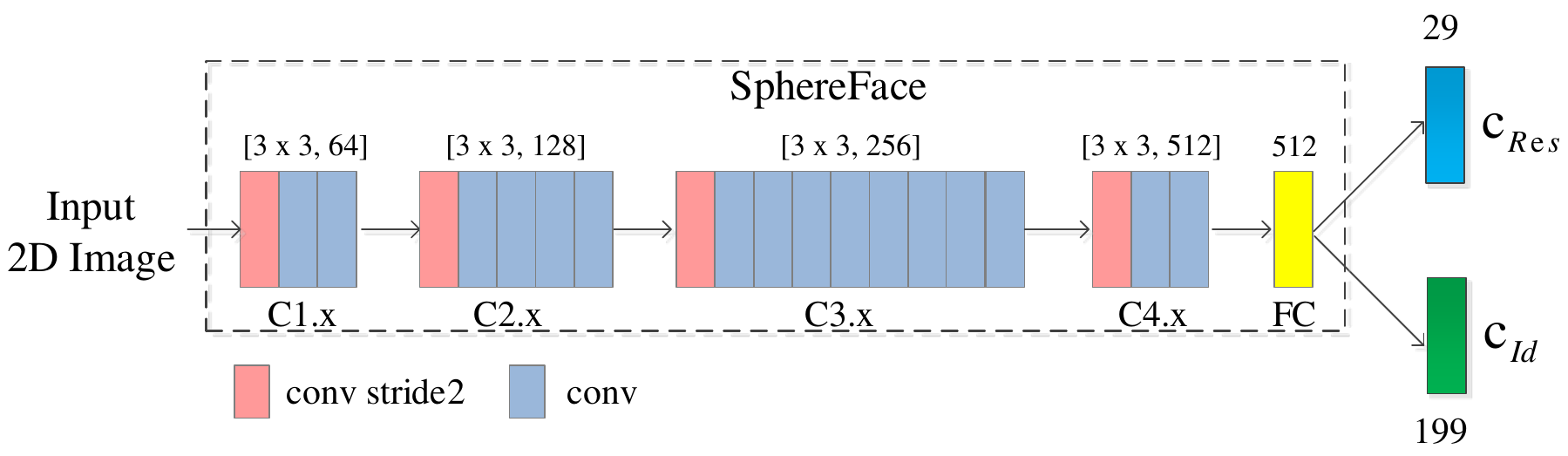}
\end{center}
   \caption{Encoder in the proposed method is implemented based on SphereFace~\cite{Liu2017SphereFace}. It converts the input $2$D image to latent identity and residual shape feature representations.}
\label{fig:encoder}
\end{figure}

We further assume that $\Delta \bold{s}_{Id}$ and $\Delta \bold{s}_{Res}$ can be described by latent representations, $\bold{c}_{Id}$ and $\bold{c}_{Res}$, respectively. This is formulated by
\begin{equation}
\Delta \bold{s}_{Id} = f_{Id}(\bold{c}_{Id}; \theta_{Id}),~\Delta \bold{s}_{Res} = f_{Res}(\bold{c}_{Res}; \theta_{Res}).
\end{equation}
Here, $f_{Id}$ ($f_{Res}$) is the mapping function that generates the corresponding shape component $\Delta \bold{s}_{Id}$ ($\Delta \bold{s}_{Res}$) from the latent representation, with parameters $\theta_{Id}$ ($\theta_{Res}$). The latent representations can be obtained from the input $2$D face image $\bold{I}$ via another function $h$:
\begin{equation}
[\bold{c}_{Id}, \bold{c}_{Res}] = h(\bold{I}; \theta),
\end{equation}
where $\theta$ are the parameters involved in $h$. Usually, the latent representations $\bold{c}_{Id}$ and $\bold{c}_{Res}$ ($\in\mathbb{R}^{Q\times 1}$) are of much lower dimension than the input $2$D face image $\bold{I}$ as well as the output $3$D face shape point cloud $\bold{s}$ (see Fig. \ref{fig:encoder}).

\subsection{An Encoder-Decoder Network}

The above composite model can be naturally implemented as an encoder-decoder network, in which $h$ serves as an encoder to extract latent representations of $2$D face images, and $f_{Id}$ and $f_{Res}$ are decoders to recover the identity and residual shape components. As shown in Fig. \ref{fig:framework}, the latent representation $\bold{c}_{Id}$ is employed as features for face recognition. In order to enhance the discriminative capability of $\bold{c}_{Id}$, we impose over $\bold{c}_{Id}$ an identification loss that can disentangle identity-sensitive from identity-irrelevant features in $3$D face shapes. Meanwhile, a reconstruction loss is applied to the $3$D face shapes generated by the decoders to guide $\bold{c}_{Res}$ and $f_{Res}$ to better capture identity-irrelevant shape components. Such an encoder-decoder network enables us to jointly learn accurate $3$D face shape reconstructor and discriminative shape features. Next, we detail the implementation of our proposed method.

\subsection{Implementation Detail}

\subsubsection{Network Structure}

\textbf{Encoder Network.} The encoder network, aiming at extracting latent identity and residual shape representations of $2$D face images, should have good capacity for discriminating different faces as well as capturing abundant detail on faces. Hence, we employ a state-of-the-art face recognition network, i.e., SphereFace~\cite{Liu2017SphereFace}, as the base encoder network. This network consists of $20$ convolutional layers and a fully-connected (FC) layer, and takes the $512$-dim output of the FC layer as the feature representation of faces. We append another two parallel FC layers to the base SphereFace network to generate $199$-dim identity latent representation and $29$-dim residual latent representation, respectively. Fig.~\ref{fig:encoder} depicts the SphereFace-based encoder network. Input $2$D face images to the encoder network are pre-processed as in~\cite{Liu2017SphereFace}: The face regions are detected by using MTCNN~\cite{Zhang2016Joint}, and then cropped and scaled to $112\times 96$ pixels whose values are normalized to the interval from $-1$ to $1$. Each dimension in the output latent representations is also normalized to the interval from $-1$ to $1$.

\begin{figure}[t]
\begin{center}
\includegraphics[width=0.85\linewidth]{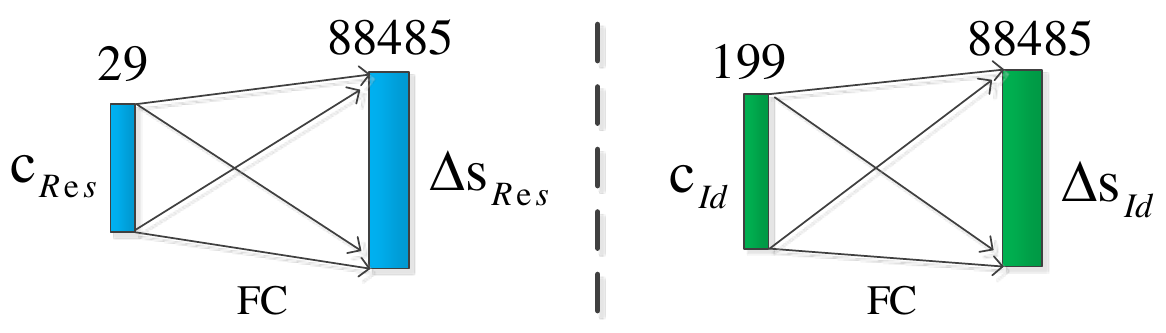}
\end{center}
   \caption{Decoders in the proposed method are implemented as a fully connected (FC) layer. They convert the latent representations to corresponding shape components. }
\label{fig:decoder}
\end{figure}

\textbf{Decoder Network.} Taking the identity and residual latent representations as input, the decoder network recovers the identity and residual shape components of $3$D face shapes. Since both the input and output of the decoder network are vectors, we use a multilayer perception (MLP) network to implement the decoder. More specifically, we use two FC layers to convert the latent representations to corresponding shape components, one for identity and the other for the residual. Fig.~\ref{fig:decoder} shows the detail of the implemented decoder network. As can be seen, the generated $3$D face point clouds have $29,495$ vertices, and the output of the MLP-based decoder network thus is $88,485$-dim. By analogy with the $3$DMM of $3$D faces, the weights of the connections between one entry in $\bold{c}_{Id}$ or $\bold{c}_{Res}$ and the output neurons can be considered as one basis of $3$DMM. Thanks to the joint training strategy, the capacity of the `bases' learnt here is much beyond that of the classical $3$DMM, as we will show in the experiments.


\textbf{Loss Functions.} We use two loss functions, $3$D shape reconstruction error and face identification error, as the supervisory signals during the end-to-end training of the encoder-decoder network. To measure the $3$D shape reconstruction error, we use the Euclidean loss, $L_{R}$, to evaluate the deviation of the reconstructed $3$D face shape from the ground truth one. The reconstructed $3$D face shape is obtained according to Eq.~(\ref{eq:shape_compose}) based on the decoder network's output $\Delta \bold{s}_{Id}$ and $\Delta \bold{s}_{Res}$ (see Fig. \ref{fig:framework}). The face identification error is measured by using the softmax loss, $L_{C}$, over the identity latent representation. The overall loss to the proposed encoder-decoder network is defined by
\begin{equation}
L = \lambda_{R}L_{R} + L_{C},
\end{equation}
where $\lambda_{R}$ is the weight for the reconstruction loss.

\subsubsection{Training Data}

To train the encoder-decoder network, we need a set of data that contain multiple $2$D face images of same subjects with their corresponding $3$D face shapes, i.e., $\{\bold{I}^{i}, l^{i}, \bold{s}^{i}\}_{i=1}^{N}$. $l^{i}\in \{1, 2, \cdots, K\}$ is the subject label of the $2$D face image $\bold{I}^{i}$ and $3$D face $\bold{s}^{i}$. $N$ is the total number of $2$D images, and $K$ is the total number of subjects in the training set. However,  such a large-scale dataset is not publicly available. Motivated by prior work~\cite{tran2016regressing}, we construct the training data from CASIA-WebFace~\cite{Yi2014Learning}, a widely-used $2$D face recognition database, via a multi-image $3$DMM fitting method, which is adapted from the method in~\cite{zhu2015high, Roth_2016_CVPR}.

Faces on the images in CASIA-WebFace are detected by using the method in~\cite{Zhang2016Joint}, and $68$ landmarks are located by the method in~\cite{bulat2017far}. We discard images where either detection or alignment fails, which results in $488,848$ images of $10,575$ different subjects in our training data. On average, each subject has $\sim 46$ images. Given the face images and their facial landmarks, we apply the following multi-image $3$DMM fitting method to estimate for each subject an identity $3$D shape component that is common to all its $2$D face images, and different residual $3$D shape components that are unique to each of the subject's $2$D images.

The $3$DMM represents a $3$D face shape as
\begin{equation}
\bold{s} = \bar{\bold{s}} + {\bold{A}}_{id}{\alpha}_{id} + {\bold{A}}_{exp}{\alpha}_{exp},
\end{equation}
where $\bold{A}_{id}$ and $\bold{A}_{exp}$ are, respectively, the identity and expression shape bases, and $\alpha_{id}$ and $\alpha_{exp}$ are the corresponding coefficients. In this paper, we use the shape bases given by the Basel Face Model \cite{paysan20093D} as $\bold{A}_{id}$, and the blendshape bases in FaceWarehouse \cite{cao2014facewarehouse} as $\bold{A}_{exp}$.

To fit the $3$DMM to $M$ images of a subject, we attempt to minimize the difference between $\bold{u}$, the landmarks detected on the images, and $\hat{\bold{u}}$, the landmarks obtained by projecting the estimated $3$D face shapes onto the images, under the constraint that all the images of the subject share the same $\alpha_{id}$. $\hat{\bold{u}}$ is computed from the estimated $3$D face shape $\hat{\bold{s}}$ (let $\hat{\bold{s}}_{U}$ denote the vertices in $\hat{\bold{s}}$ corresponding to the landmarks) by $\hat{\bold{u}} = f\cdot \bold{P}\cdot \bold{R}\cdot (\hat{\bold{s}}_{U} + \bold{t})$, where $f$ is the scale factor, $\bold{P}$ is the orthographic projection, $\bold{R}$ and $\bold{t}$ are the rotation matrix and translation vector in $3$D space. Mathematically, our multi image $3$DMM fitting optimizes the following objective:
\begin{equation}\label{eq:multi3Dmmfitting}
\mathop{\min}_{\alpha_{id}, \{f^{j}, \bold{R}^{j}, \bold{t}^{j}, \alpha_{exp}^{j}\}_{j=1}^{M}} \ \\ \sum_{j=1}^{M} \|\bold{u}^{j}-\hat{\bold{u}}^{j}\|_{2}^{2}.
\end{equation}

We solve the optimization problem in Eq.~(\ref{eq:multi3Dmmfitting}) in an alternating way. As an initialization, we set both $\alpha_{id}$ and $\alpha_{exp}$ to zero. We first estimate the projection parameters $\{f^{j}, \bold{R}^{j}, \bold{t}^{j}\}_{j=1}^{M}$, then expression parameters $\{\alpha_{exp}^{j}\}_{j=1}^{j=M}$, and lastly identity parameters $\alpha_{id}$. When estimating one of the three sets of parameters, the rest two sets of parameters are fixed as they are. The optimization is repeated until the objective function value does not change. We have typically found this to converge within seven iterations.

\subsubsection{Training Process}

With the prepared training data, we train our encoder-decoder network in three phases. In Phase I, we train the encoder by setting the target latent representations as $\bold{c}_{Id} = \alpha_{id}$ and $\bold{c}_{Res} = \alpha_{exp}$ and using Euclidean loss. In Phase II, we train the decoder for the identity and residual components separately. In Phase III, the end-to-end joint training is conducted based on the pre-trained encoder and decoder. Considering that the network already has good performance in reconstruction after pre-training, we first lay more emphasis on recognition in the joint loss function by setting $\lambda_{R}$ to $0.5$. When the loss function gets saturated (usually within $10$ epochs), we continue the training by updating $\lambda_{R}$ to $1.0$. The joint training concludes in about another $20$ epochs.

It is worth mentioning that the recovered $3$DMM parameters are directly used as the latent representations during pre-training. This provides a good initialization for the encoder-decoder network, but limits the network to the capacity of the pre-determined $3$DMM bases. The joint training in Phase III alleviates such limitation by utilizing the identification loss as a complementary supervisory signal to the reconstruction loss. As a result, the learnt encoder-decoder network can better disentangle identity from non-identity information in $3$D face shapes, and thus enhance face recognition accuracy without impairing the $3$D face reconstruction accuracy.


\section{Experiments}

Two sets of experiments have been done to evaluate the effectiveness of the proposed method in $3$D face reconstruction and face recognition. The MICC~\cite{bagdanov2011florence} and BU$3$DFE~\cite{yin20063D} databases are used for experiments of $3$D face reconstruction, and the LFW~\cite{huang2007labeled} and YTF~\cite{Wolf2011Face} databases are used in face recognition experiments. Next, we report the experimental results~\footnote{More experimental results are provided in the supplementary material.}.

\subsection{$3$D Shape Reconstruction Accuracy}

\begin{table*}[t]
\footnotesize
\centering
\renewcommand\arraystretch{0.75}
\caption{$3$D face reconstruction accuracy (RMSE) under different yaw angles on the BU$3$DFE database.}
\begin{tabular*}{13.7cm}{l c c c c c c c c c c c}
\toprule
Method & $\pm90^{\circ}$ & $\pm80^{\circ}$ & $\pm70^{\circ}$ & $\pm60^{\circ}$ & $\pm50^{\circ}$ & $\pm40^{\circ}$ & $\pm30^{\circ}$ & $\pm20^{\circ}$ & $\pm10^{\circ}$ & $0^{\circ}$ & Avg.\\ 
\midrule 
\midrule 
 VRN  & $6.96$ & $6.20$ & $6.14$ & $6.01$ & $5.91$ & $5.50$ & $4.93$ & $3.86$ & $3.70$ & $3.66$ & $5.29$\\
 $3$DDFA & $2.90$ & $2.88$ & $2.81$ & $2.82$ & $2.77$ & $2.79$ & $2.76$ & $2.73$ & $2.55$ & $2.48$ & $2.75$\\
 $3$DMM-CNN  & - & - & - & - & $2.30$ & $2.26$ & $2.23$ & $2.22$ & $2.19$ & $2.17$ & $2.23$\\
 $3$DSR & $2.11$ & $2.11$ & $2.12$ & $2.13$ & $2.16$ & $2.14$ & $2.12$ & $2.10$ & $2.10$ & $2.09$ & $2.12$ \\
\midrule 
Proposed &  $\bf{2.09}$ & $\bf{2.04}$ &  $\bf{2.03}$ &  $\bf{2.03}$ &  $\bf{2.00}$ &  $\bf{1.99}$ &  $\bf{2.03}$ &  $\bf{2.01}$  & $\bf{1.97}$  &  $\bf{1.93}$ & $\bf{2.01}$ \\
\bottomrule 
\end{tabular*}
\label{tab:bu3D_pose_compare}
\end{table*}


The $3$D face reconstruction accuracy is assessed by using $3$D Root Mean Square Error (RMSE)~\cite{tran2016regressing}, defined as $\texttt{RMSE} = \frac{1}{N_{T}}\sum_{i=1}^{N_{T}}(\| \bold{s}^{*}_{i}-\hat{\bold{s}}_{i} \|/n),$ where $N_{T}$ is the total number of testing samples, $\bold{s}^{*}_{i}$ and $\hat{\bold{s}}_{i}$ are the ground truth and reconstructed $3$D face shape of the $i^{\texttt{th}}$ testing sample. To compute the RMSE, the reconstructed $3$D faces are first aligned to ground truth via Procrustes global alignment based on $68$ $3$D landmarks as suggested by~\cite{bas2016fitting}, and then cropped at a radius of $95mm$ around the nose tip.

We compare our method with four state-of-the-art $3$D face reconstruction methods, $3$DDFA~\cite{zhu2015high}, $3$DMM-CNN~\cite{tran2016regressing}, $3$D shape regression based ($3$DSR) method~\cite{liu2016joint}, and VRN~\cite{jackson2017vrn}. 
Among them, the first two methods reconstruct $3$D face shapes via estimating $3$DMM parameters, while the other two directly regress $3$D face shapes  from either landmarks or  $2$D images. $3$DMM-CNN method is the only existing method that takes into consideration the discriminative power of the estimated $3$DMM parameters. $3$DSR method generates pose and expression normalized $3$D face shapes that are believed to be more beneficial to face recognition. For those methods that need facial landmarks on $2$D images, we use the method in~\cite{bulat2017far} to automatically detect the landmarks.

\begin{figure}[t]
\centering
\includegraphics[width=\linewidth]{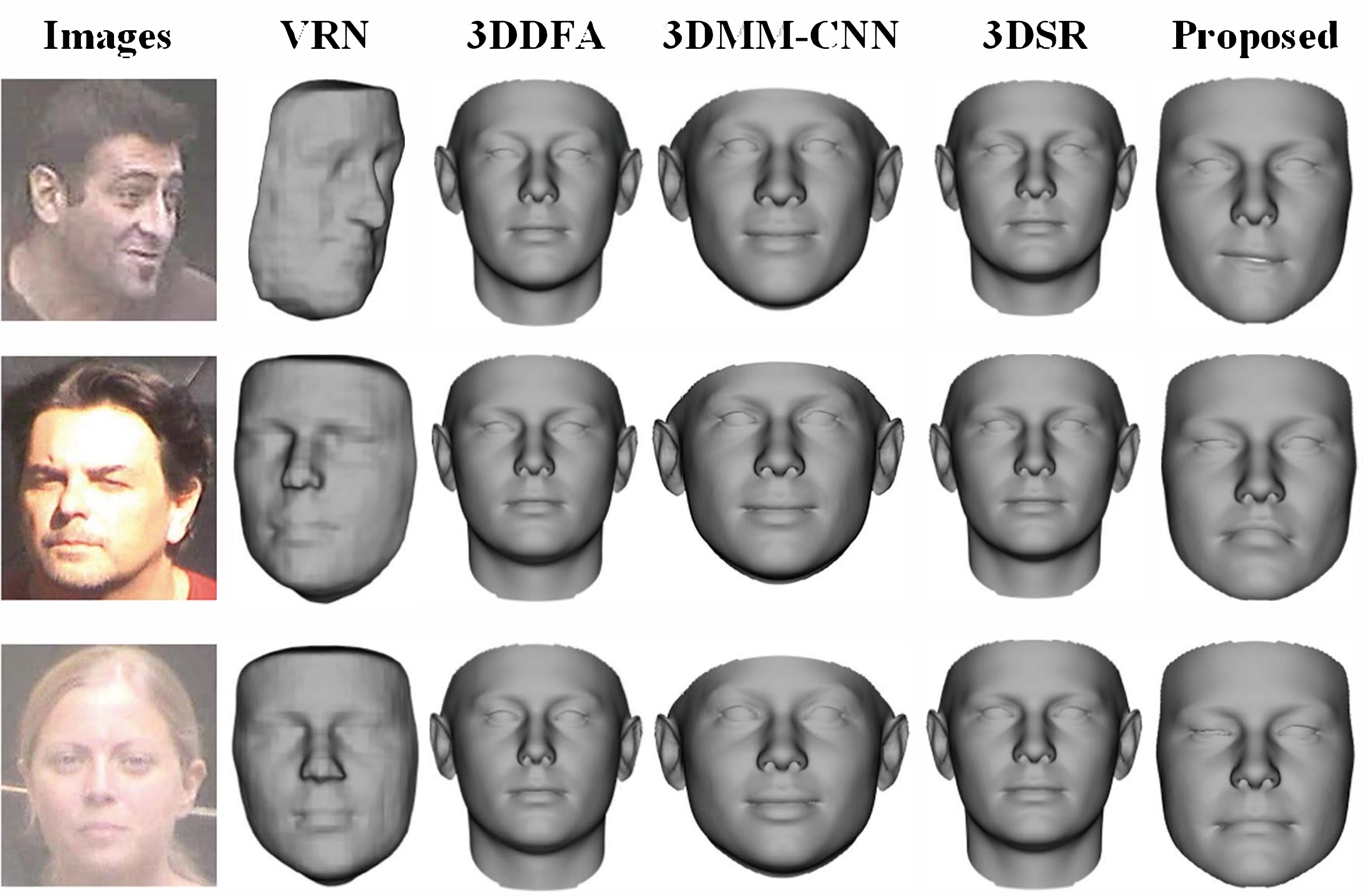}
\caption{Reconstruction results for three MICC subjects. The first column shows the input images, and the rest columns show the reconstructed $3$D shapes that have the same expression as the input images, using the methods of VRN \cite{jackson2017vrn}, $3$DDFA \cite{zhu2015high}, $3$DMM-CNN\cite{tran2016regressing}, $3$DSR \cite{liu2016joint} and the proposed method.}
\label{fig:micc_example}
\end{figure}

\textbf{Results on MICC.} The MICC database contains three challenging face videos and ground-truth $3$D models acquired using a structured-light scanning system for each of $53$ subjects. The videos span the range of controlled indoor to unconstrained outdoor settings. 
The outdoor videos are very challenging due to the uncontrolled lighting conditions. In this experiment, we randomly select $5,000$ images from $31,466$ outdoor video frames of $53$ subjects. Table~\ref{tab:micc_compare} shows the $3$D face reconstruction error of different methods on the MICC database. As can be seen, our proposed method obtains the best accuracy due to its fine-grained processing of features in $3$D face shapes. Note that VRN, the first method in the literature that regresses $3$D face shapes directly from $2$D images, has relatively high reconstruction error in terms of RMSE, mainly because it generates low-resolution $3$D face shapes as volumetric representations. In contrast, we reconstruct high-resolution (dense) $3$D face shapes as point clouds with help from low dimensional latent representations.

\begin{table}[t]
\footnotesize
\renewcommand\arraystretch{0.75}
\centering
\caption{$3$D face reconstruction accuracy on the MICC database.}
\begin{tabular*}{8.3cm}{l | p{0.8cm}<{\centering} p{0.8cm}<{\centering} p{1.6cm}<{\centering} p{0.8cm}<{\centering} p{1cm}<{\centering}}
\toprule
Method & VRN & $3$DDFA & $3$DMM-CNN & $3$DSR & Proposed\\ 
\midrule 
\midrule 
RMSE & $5.34$ & $2.73$ & $2.20$ & $2.07$ & $\bf{2.00}$  \\
\bottomrule 
\end{tabular*}
\label{tab:micc_compare}
\end{table}

\textbf{Results on BU$3$DFE.} The BU$3$DFE database contains $3$D faces of $100$ subjects displaying expression of neutral (NE), happiness (HA), disgust (DI), fear (FE), anger (AN), surprise (SU) and sadness (SA). 
All non-neutral expressions were acquired at four levels of intensity. 
We select neutral and the first intensity level of the rest six expressions as testing data, resulting in $700$ testing samples. Further, we render another set of testing images of neutral expression at different poses, i.e., $-90^{\circ}$ to $90^{\circ}$ yaws with a $10^{\circ}$ interval. These two testing sets evaluate the reconstruction  across expressions and poses, respectively.

%
%
%
%
%
\begin{table*}[t]
\centering
\caption{Face recognition accuracy on the LFW and YTF databases.}
\renewcommand\arraystretch{0.82}
\footnotesize
\begin{tabular*}{15.1cm}{l c c c c c c c}
\toprule
Method & Shape  & Texture & Accuracy & 100\%-EER & AUC & TAR-10\% & TAR-1\% \\
\midrule 
\midrule 
\multicolumn{8}{c}{\textbf{Labeled Faces in the Wild (LFW)}}\\
\toprule
\multirow{3}{*}{$3$DMM}    & ${\surd}$ &  \textbf{$\times$}  &$66.13\pm2.79$  &$65.70\pm2.81$  &$72.24\pm2.75$  &$35.90\pm3.74$  &$12.37\pm4.81$ \\
                                 & \textbf{$\times$} &   ${\surd}$ &$74.93\pm1.14$  &$74.50\pm1.21$  &$82.94\pm1.14$  &$60.40\pm3.15$  &$28.73\pm7.17$  \\
                                 & ${\surd}$ &        ${\surd}$    &$75.25\pm2.12$  &$74.73\pm2.56$  &$83.21\pm1.93$  &$59.40\pm4.64$  &$29.67\pm4.73$ \\
\cline{1-1}
$3$DDFA   & ${\surd}$ &  \textbf{$\times$}                   &$66.98\pm2.56$  &$67.13\pm1.90$  &$73.30\pm2.49$  &$36.76\pm6.27$  &$10.00\pm3.22$ \\
\cline{1-1}
\multirow{3}{*}{$3$DMM-CNN}& ${\surd}$ &  \textbf{$\times$}  &$90.53\pm1.34$  &$90.63\pm1.61$  &$96.60\pm0.79$  &$91.13\pm2.62$  &$58.20\pm12.14$ \\
                                 & \textbf{$\times$} &   ${\surd}$ &$90.60\pm1.07$  &$90.70\pm1.17$  &$96.75\pm0.59$  &$91.23\pm2.42$  &$52.60\pm8.14$ \\
                                 & ${\surd}$ &        ${\surd}$    &$92.35\pm1.29$  &$92.33\pm1.33$  &$97.71\pm0.64$  &$94.20\pm2.00$  &$65.57\pm6.93$ \\
\hline
Proposed   & ${\surd}$ &  \textbf{$\times$}                   &$\bf{94.43\pm1.47}$  & $\bf{94.40\pm1.52}$   &$\bf{98.12\pm0.90}$  &$\bf{95.07\pm2.39}$  &$\bf{74.54\pm4.33}$ \\

\midrule 
\midrule 

\multicolumn{8}{c}{\textbf{YouTube Faces (YTF)}}\\
\toprule
\multirow{3}{*}{$3$DMM}    & ${\surd}$ &  \textbf{$\times$}  &$73.26\pm2.51$  &$73.08\pm2.65$  &$80.41\pm2.60$  &$51.36\pm5.11$  &$24.04\pm4.56$ \\
                                 & \textbf{$\times$} &   ${\surd}$ &$77.34\pm2.54$  &$76.96\pm2.64$  &$85.32\pm2.63$  &$63.16\pm5.07$  &$31.36\pm5.21$  \\
                                 & ${\surd}$ &        ${\surd}$    &$79.56\pm2.08$  &$79.20\pm2.07$  &$87.35\pm1.92$  &$69.08\pm5.00$  &$34.56\pm6.89$ \\
\cline{1-1}
$3$DDFA   & ${\surd}$ &  \textbf{$\times$}                   &$68.10\pm2.93$  &$67.96\pm3.12$  &$74.95\pm3.04$  &$40.52\pm3.65$  &$12.20\pm2.67$ \\
\cline{1-1}
\multirow{3}{*}{$3$DMM-CNN}& ${\surd}$ &  \textbf{$\times$}  &$88.28\pm1.84$  &$88.32\pm2.16$  &$95.95\pm1.38$  &$86.60\pm3.95$  &$51.12\pm8.86$ \\
                                 & \textbf{$\times$} &   ${\surd}$ &$87.56\pm2.56$  &$87.68\pm2.25$  &$94.44\pm1.38$  &$84.80\pm4.89$  &$40.92\pm8.26$ \\
                                 & ${\surd}$ &        ${\surd}$    &$\bf{88.80\pm2.21}$  &$\bf{88.84\pm2.40}$  &$95.37\pm1.43$  &$87.92\pm4.18$  &$46.56\pm6.20$ \\
\hline
Proposed   & ${\surd}$ &  \textbf{$\times$}                   &$88.74\pm1.03$  &$88.70\pm1.15$ &$\bf{96.28\pm0.63}$  &$\bf{89.00\pm2.40}$  &$\bf{53.44\pm4.51}$ \\

\bottomrule 
\end{tabular*}
\label{tab:verification}
\end{table*}

\begin{figure}[t]
\centering
\includegraphics[width=0.79\linewidth]{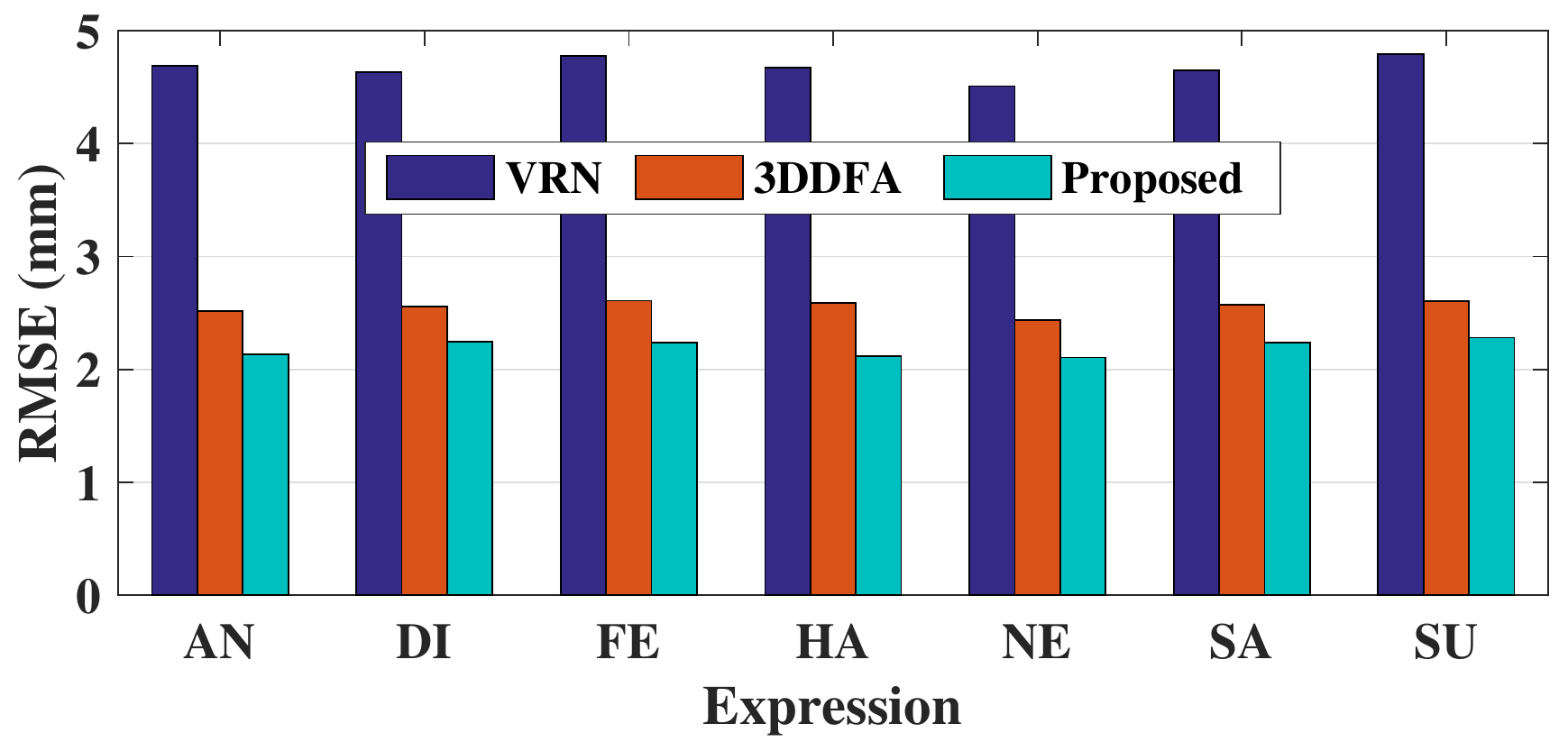}
\caption{Reconstruction accuracy of $3$D face shapes under different expressions on the BU$3$DFE database. The mean RMSEs of thee methods over all expressions are $4.68$, $2.56$, and $2.19$ respectively.}
\label{fig:bu3D_exp_error}
\end{figure}

\begin{figure}[t]
\centering
\includegraphics[width=0.79\linewidth]{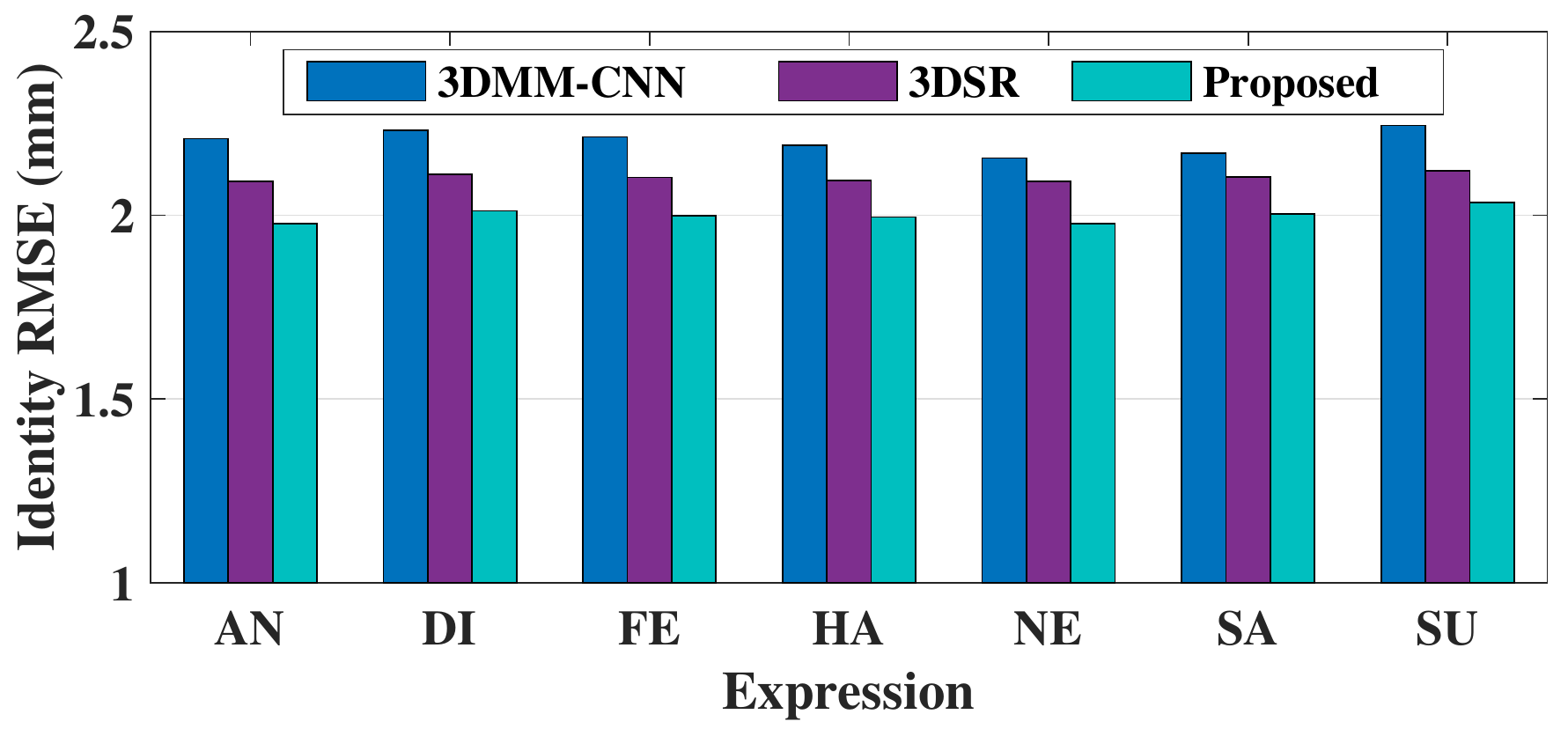}
\caption{Reconstruction accuracy of the identity component of $3$D face shapes under different expressions on the BU$3$DFE database. The mean RMSEs of thee methods over all expressions are $2.21$, $2.10$, and $2.00$ respectively.}
\label{fig:bu3D_neu_error}
\end{figure}

Table~\ref{tab:bu3D_pose_compare} shows the \emph{reconstruction error across poses} (i.e., yaw) of different methods. 
It can be seen that the RMSE of the proposed method is lower than that of baselines. Moreover, as the pose angle becomes large, the error of our method does not increase substantially. 
This proves the robustness of the proposed method to pose variations. Figure~\ref{fig:bu3D_exp_error} shows the \emph{reconstruction error across expressions} of VRN, $3$DDFA, and the proposed method based on their reconstructed $3$D face shapes that have the same expression as the input images. Figure~\ref{fig:bu3D_neu_error} compares $3$DMM-CNN, $3$DSR, and the proposed method in terms of RMSE of their reconstructed identity or expression-normalized $3$D face shapes. 
These results demonstrate the superiority of the proposed method over baselines in handling  expressions.

\begin{figure}
\centering
\includegraphics[trim={7mm 0 0 0},clip,width=1\linewidth]{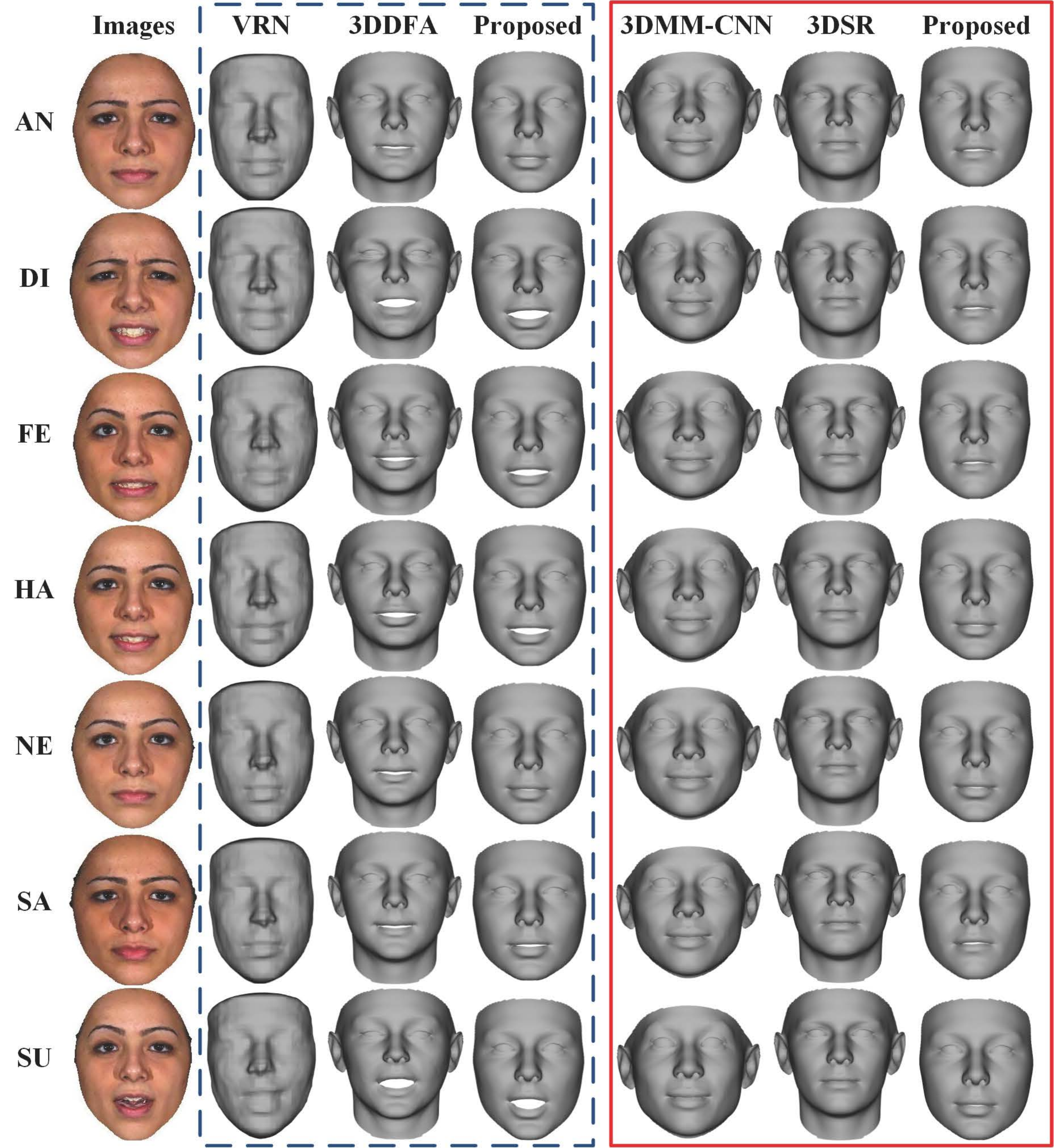}
\caption{Reconstruction results for an BU$3$DFE subject under seven different expressions. The first column shows the input images. In the blue box, we show the reconstructed $3$D shapes that have the same expression as the input images, using the methods of VRN~\cite{jackson2017vrn}, $3$DDFA~\cite{zhu2015high} and the proposed method. In the red box, we show the reconstructed \emph{identity} $3$D shapes obtained by $3$DMM-CNN~\cite{tran2016regressing}, $3$DSR~\cite{liu2016joint} and the proposed method. Our composite 3D shape model enables us to generate two types of 3D shapes.}
\label{fig:bu3D_exp_example}
\end{figure}

Some example $3$D face reconstruction results are shown in Fig.~\ref{fig:micc_example} and Fig.~\ref{fig:bu3D_exp_example}. From these results, we can clearly see that the proposed method not only performs well in reconstructing accurate $3$D face shapes for in-the-wild $2$D images, but also disentangles identity and non-identity (e.g., expression) components in $3$D face shapes. As we will show in the following face recognition experiments, the disentangled shape features contribute to face recognition.

\begin{figure*}[t]
\centering
\includegraphics[width=0.98\linewidth]{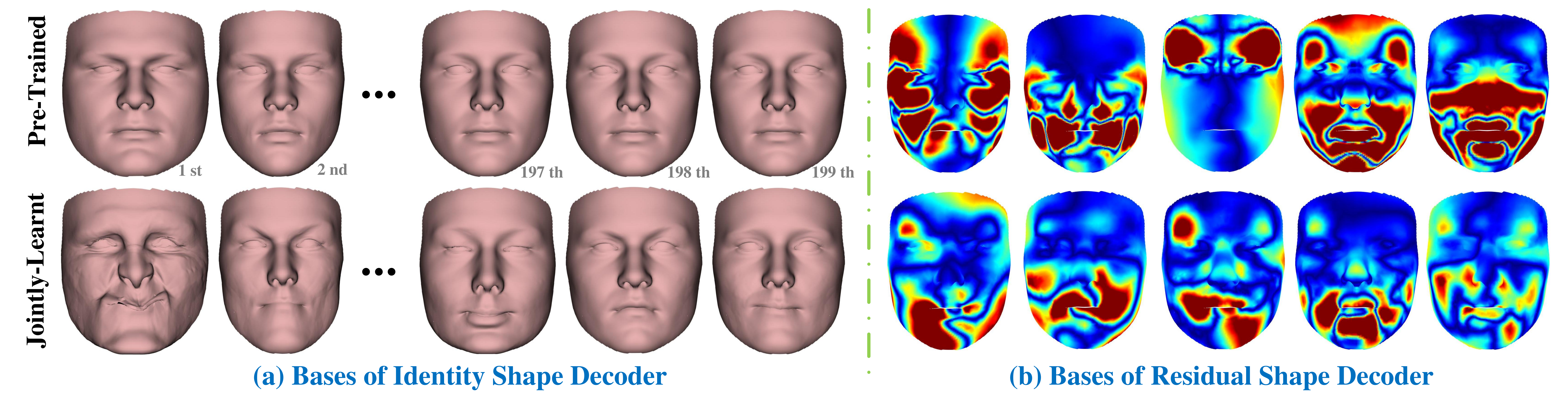}
\caption{Comparing the pre-trained 3DMM-like and our jointly-learnt bases defined by the weights of identity and residual shape decoders. (a) For the bases of identity shape decoder, the weights associated with each entry in $\bold{c}_{Id}$ are added to the mean shape, reshaped to a point cloud ($\in \mathbb{R}^{3\times n}$), and shown as polygon meshes. (b) For the bases of residual shape decoder, the weights associated with each entry in $\bold{c}_{Res}$ are reshaped to a point cloud ($\in \mathbb{R}^{3\times n}$), and shown as a heat map that measures the norm value of each vertex (i.e., the deviation from the identity shape). Red colors in the heat maps indicate larger deviations. It is important to note that the conventional $3$DMM bases are trained from $3$D face scans, while our bases are learnt from $2$D images.}
\label{fig:visualization}
\end{figure*}

\subsection{Face Recognition Accuracy}

To evaluate the effectiveness of our shape features (i.e., the identity representations) to face recognition, we compute the similarity of two faces using the cosine distances between their shape features extracted by the encoder of our method. 
To investigate the complementarity between our learnt shape features and existing texture features, we also fuse our method with existing methods via summation at the score level~\cite{kittler1998combining}. 
The counterpart methods we consider here include $3$DMM~\cite{romdhani2005estimating}, $3$DDFA~\cite{zhu2015high}, $3$DMM-CNN~\cite{tran2016regressing}, and SphereFace~\cite{Liu2017SphereFace}. 
We compare the methods in terms of verification accuracy, $100\%$-EER (Equal Error Rate), AUC (Area Under Curve) of ROC (Receiver Operating Characteristic) curves, and TAR (True Acceptance Rate) at FAR (False Acceptance Rate) of $10\%$ and $1\%$.

\textbf{Results on LFW.} The Labeled Faces in the Wild (LFW) benchmark dataset contains $13,323$ images collected from Internet. 
The verification set consists of $10$ folders, each with $300$ same-person pairs and $300$ different-person pairs. 
The recognition accuracy of different methods on LFW is listed in Tab.~\ref{tab:verification}. Among all the $3$D face reconstruction methods, when using only shape features, our proposed method achieves the highest accuracy, improving TAR@$1\%$ FAR from $58.20\%$ to $74.54\%$ with respect to the latest $3$DMM-based method~\cite{tran2016regressing}.

\textbf{Results on YTF.} The YouTube Faces (YTF) database contains $3,425$ videos of $1,595$ individuals. Face images (video frames) in YTF have lower quality than those in LFW, due to larger variations in pose, illumination and expression, and low resolution as well. Table~\ref{tab:verification} summarizes the recognition accuracy of different methods on YTF. Despite the low-quality face images, our proposed method still outperforms the baseline methods in the sense of extracting discriminative shape features. By fusing with one of the state-of-the-art texture-based face recognition methods (i.e., SphereFace~\cite{Liu2017SphereFace}), our proposed method further improves the face recognition accuracy on YTF from $94.78\%$ to $95.18\%$. This proves the complementarity of \emph{properly reconstructed} shape features to texture features in face recognition. 
This is a notable result especially considering the $2$D face recognition method of SphereFace~\cite{Liu2017SphereFace} has already set a very high baseline (i.e., $94.78\%$).



\begin{table}
\footnotesize
\renewcommand\arraystretch{0.8}
\centering
\caption{Efficiency comparison of different methods.}
\begin{tabular*}{8.3cm}{l | p{0.6cm}<{\centering} p{0.8cm}<{\centering} p{1.6cm}<{\centering} p{0.6cm}<{\centering} p{1cm}<{\centering}}
\toprule
Method & VRN & $3$DDFA & $3$DMM-CNN & $3$DSR & Proposed\\ 
\midrule 
\midrule 
Time (ms) & $55.68$  & $39.17$ & $30.12$ & $29.80$ & $\bf{4.79}$  \\
\bottomrule 
\end{tabular*}
\label{tab:runtime}
\end{table}

\subsection{Computational Efficiency}
To assess the computational efficiency, we run the methods on a PC (with an Intel Core i$7$-$5930$K @ $3.5$GHz, $32$GB RAM and an GeForce GTX $1080$) for $700$ images, and calculate the average runtime per image in Tab.~\ref{tab:runtime}.
Note that $3$DDFA and $3$DMM-CNN estimate the $3$DMM parameters in the first step, and we report their runtime of obtaining the final $3$D faces.
For VRN, $3$DDFA and $3$DMM-CNN, despite stand-alone landmark detection is required, the reported time does not include the landmark detection time.
Our proposed method needs only $4.79$ milliseconds (ms) per image, which is an order of magnitude faster than baseline methods.
This is owing to the light-weight network in our method.
In contrast, baseline methods use either very deep networks \cite{tran2016regressing}, or cascade approaches \cite{Richardson2016Learning, liu2016joint}.

\subsection{Analysis and Discussion}
To offer insights into the learnt decoders, we visualize their weight parameters in Fig.~\ref{fig:visualization}. The weights associating one entry in the latent representations with all the neurons in the FC layer in the decoders are analogous to a $3$DMM basis (see Fig.~\ref{fig:decoder}). Both pre-trained bases and jointly-learnt bases are shown for comparison in Fig.~\ref{fig:visualization}, from which the following observations can be made.

(i) The pre-trained identity bases approximate the conventional $3$DMM bases~\cite{blanz1999morphable} that are ordered with latter bases capturing less shape variations. In contrast, our jointly-learnt identity bases all describe rich shape variations.

(ii) Some basis shapes in the jointly-learnt bases do not look like regular face shapes. We believe this is due to the employed joint reconstruction and identification loss function. The bases trained from a set of $3$D scans as in $3$DMM, while optimal for reconstruction,  might limit the discriminativeness of shape parameters. Our bases are trained with the classification in mind, which ensures the superior performance of our method in face recognition.

(iii) The pre-trained residual bases, like the expression shape bases~\cite{cao2014facewarehouse}, appear symmetrical. The jointly-learnt residual bases display more diverse shape deviation patterns. 
This indicates that the residual shape deformation captured by the jointly-learnt bases is much beyond that caused by expression changes, and proves the effectiveness of our method in disentangling $3$D face shape features.



\section{Conclusions}
We have proposed a novel encoder-decoder-based method for jointly learning discriminative shape features from a $2$D face image and reconstructing its dense $3$D face shape. To train the encoder-decoder network, we implement a multi-image $3$DMM fitting method to construct training data, and develop an effective training scheme with a joint reconstruction and identification loss. We show with comprehensive experimental results that the proposed method can effectively disentangle identity and non-identity features in $3$D face shapes and thus achieve state-of-the-art $3$D face reconstruction accuracy as well as improved face recognition accuracy. 



\section*{Supplementary Material}

In this supplementary material, we provide additional experimental results, including
\begin{itemize}
\item[-] Face recognition results on IJB-A database;
\item[-] Phase-by-Phase Evaluation: CNN vs. 3DMM;
\item[-] Qualitative reconstruction results.
\end{itemize}


\subsection*{Recognition Results on IJB-A}
The IJB-A database \cite{Klare2015Pushing}, including 5,396 images and 20,412 video frames of 500 subjects, has full pose variation and is more challenging than LFW \cite{huang2007labeled}. We evaluate both face verification (1:1 comparison) and face identification (1:N search) performance of our proposed method with comparison to existing methods on the IJB-A database. The faces are firstly automatically detected by using the method in \cite{Zhang2016Joint} and aligned by the method in \cite{bulat2017far}. If the automated methods fail, we manually crop the faces. The results are reported in Table~\ref{tab:ijba_results}.

\begin{table*}[t]
\centering
\caption{Face verification and identification performance on the IJB-A database.}
\renewcommand\arraystretch{0.9}
\small
\begin{tabular*}{13.1cm}{l c c c c c c }
\toprule
Method & Shape  & Texture & TAR-10\% & TAR-1\% & Rank-1 & Rank-5  \\
\midrule 
\midrule 
\multirow{3}{*}{3DMM}    & ${\surd}$ &  \textbf{$\times$}  &$60.7\pm2.0$  &$30.6\pm3.2$  &$34.3\pm2.2$  &$55.1\pm2.1$   \\
                                 & \textbf{$\times$} &   ${\surd}$ &$71.1\pm1.8$  &$39.5\pm4.8$  &$49.8\pm2.5$  &$69.5\pm1.4$    \\
                                 & ${\surd}$ &        ${\surd}$    &$75.4\pm1.6$  &$46.6\pm5.1$  &$57.2\pm1.9$  &$74.4\pm1.3$   \\
\cline{1-1}
3DDFA    & ${\surd}$ &  \textbf{$\times$}  &$43.3\pm2.5$  &$12.5\pm1.9$  &$16.7\pm1.9$  &$38.3\pm2.7$   \\
\cline{1-1}
\multirow{3}{*}{3DMM-CNN}& ${\surd}$ &  \textbf{$\times$}  &$86.0\pm1.7$  &$55.9\pm5.5$  &$72.3\pm1.4$  &$88.0\pm1.4$   \\
                         & \textbf{$\times$} &   ${\surd}$ &$83.5\pm2.2$  &$50.3\pm5.8$  &$70.9\pm1.5$  &$87.3\pm1.1$   \\
                         & ${\surd}$ &        ${\surd}$    &$87.0\pm1.5$  &$60.0\pm5.6$  &$76.2\pm1.8$  &$89.7\pm1.0$   \\
\hline
DRGAN    & \textbf{$\times$}  &  ${\surd}$   & $\bf{-}$     &$75.5\pm2.8$  &$84.3\pm1.3$  &$93.2\pm0.8$    \\
\cline{1-1}
Proposed   & ${\surd}$ &  \textbf{$\times$}                   &$\bf{89.6\pm1.2}$  & $58.8\pm4.9$   & $75.7\pm1.9$ & $88.2\pm1.1$ \\
\cline{1-1}
DRGAN+Proposed   & ${\surd}$  &  ${\surd}$ &$\bf{-}$  &  $\bf{76.5\pm4.2}$   & $\bf{85.4\pm1.8}$ & $\bf{93.9\pm0.9}$  \\

\bottomrule 
\end{tabular*}
\label{tab:ijba_results}
\end{table*}

\begin{table*}[t]
\centering
\renewcommand\arraystretch{0.9}
\small
\newcommand{\tabincell}[2]{\begin{tabular}{@{}#1@{}}#2\end{tabular}}
\caption{Reconstruction and recognition accuracy on different test data sets when identity disentangling and identification loss are used or not used. Refer to the paper for test data set details.}
\begin{tabular*}{17.4cm}{p{1.2cm}<{\centering} | p{1.65cm}<{\centering} | p{1.6cm}<{\centering} | c   c   c | c  c}
\toprule
\multirow{2}{*}{\tabincell{l}{Training \\ Phase}} & \multirow{2}{*}{\tabincell{l}{Identity \\ Disentangling}} & \multirow{2}{*}{\tabincell{l}{Identification \\ Loss}}  & \multicolumn{3}{c|}{Reconstruction RMSE on} & \multicolumn{2}{c}{Recognition Accuracy on}\\
\cline{4-6} \cline{7-8}
& & & MICC & BU3DFE (pose) & BU3DFE (exp.) & LFW & YTF\\
\midrule 
\midrule 

\textbf{--}  &  \textbf{$\times$}  &  \textbf{$\times$} & $2.51\pm0.57$       &  $2.54\pm0.67$      &  $2.62\pm0.73$     & \textbf{--} & \textbf{--}   \\
\hline
\textbf{II} &     ${\surd}$       &  \textbf{$\times$} & $2.23\pm0.48$       &  $2.31\pm0.55$      &  $2.45\pm0.62$      &  $68.00\pm2.21$ & $69.19\pm1.91$\\
\hline
\textbf{III}&     ${\surd}$       &  ${\surd}$         & $\bf{2.00\pm0.32}$  &  $\bf{2.01\pm0.49}$ &  $\bf{2.19\pm0.54}$ & $\bf{94.43\pm1.47}$ & $\bf{88.74\pm1.03}$\\

\bottomrule 
\end{tabular*}
\label{tab:visualization_compare}
\end{table*}

When using only reconstructed shape features, our proposed method obtains the best face recognition accuracy in terms of true acceptance rate at false acceptance rate of $10\%$ (TAR-$10\%$) and $1\%$ (TAR-$1\%$), and rank-1 and rank-5 identification rate. Although it is outperformed by DR-GAN \cite{LuanPose2017}, a state-of-the-art texture-based face recognition method, the face recognition accuracy can be further improved after combining them by score-level summation fusion. These results, consistent with the results on the LFW and YTF \cite{Wolf2011Face} databases, prove the effectiveness of our proposed method in disentangling discriminative shape features that are complementary to texture features in face recognition as well as in surpassing the conventional 3D morphable model (3DMM) bases \cite{blanz2003face} in capturing facial detail. 

\begin{figure}[t]
\begin{center}
   \includegraphics[width=0.96\linewidth]{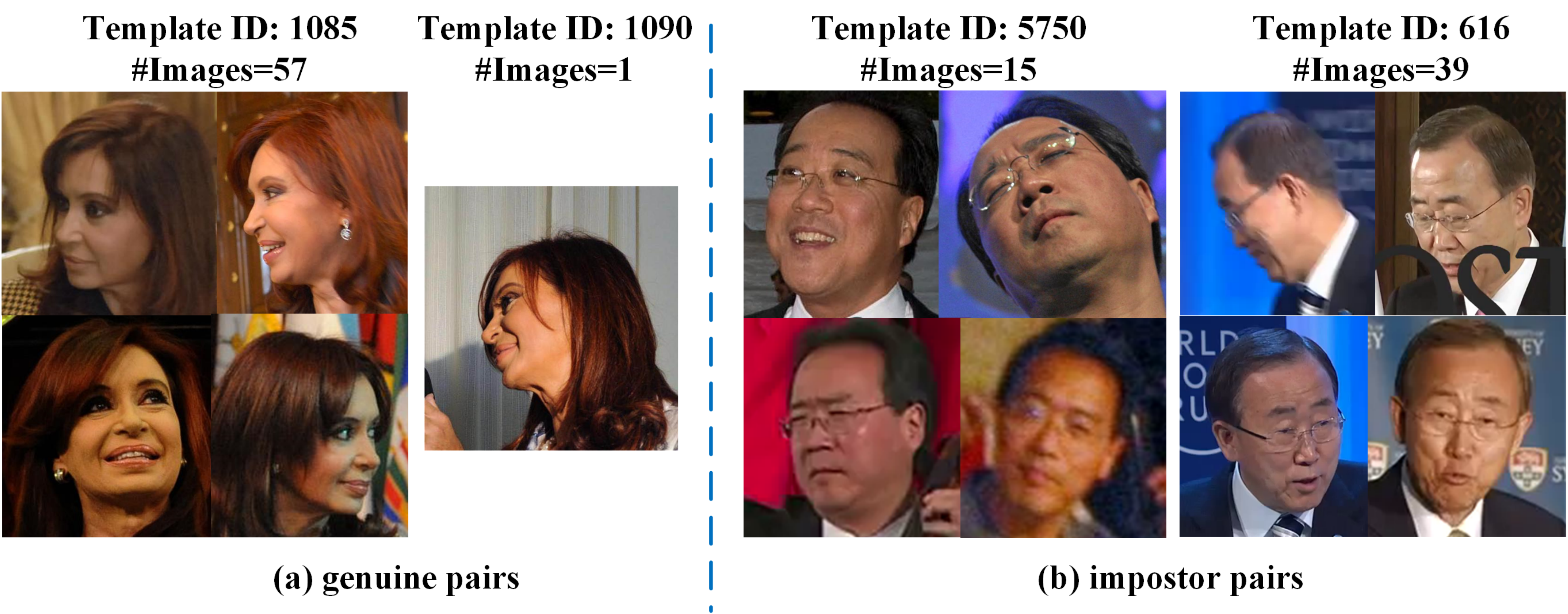}
\end{center}
   \caption{Example (a) genuine pairs and (b) imposter pairs in IJB-A, for which the state-of-the-art texture-based face recognition method (i.e., DR-GAN \cite{LuanPose2017}) fails, whereas its fusion with our proposed method succeeds.}
   \label{fig:example}
\end{figure}

Figure~\ref{fig:example} shows some example genuine and imposter pairs in IJB-A, which are incorrectly recognized by DR-GAN \cite{LuanPose2017}, but correctly recognized by the fusion of DR-GAN and our proposed method. As can be seen, while extremely large head rotations may lead to the failure of existing texture-based face recognition methods, our proposed method explores complementary shape features to robustly recognize the off-angle faces with large rotations.

\subsection*{Phase-by-Phase Evaluation: \small{CNN vs. 3DMM}}

Our proposed model is trained in three phases. Phases I and II replicate 3DMM for a proper initialization of our model, while Phase III makes our model beyond 3DMM by using joint supervisory of reconstruction and recognition (i.e., both reconstruction loss and identification loss). To address the reviewer’s concern, we compare the reconstruction and recognition results at different training phases. Table~\ref{tab:visualization_compare} gives the reconstruction results at Phases II and III, and summarizes the recognition results. It can be seen that reconstruction errors are further reduced after incorporating identification loss in Phase III. As for recognition, the accuracy is significantly improved from Phase II to Phase III. This reveals the limited discrimination power of 3DMM representations and the importance of CNN-based joint learning in expanding the representation and discrimination capacity of 3DMM-like bases.


\begin{figure*}[t]
\begin{center}
   \includegraphics[width=0.95\linewidth]{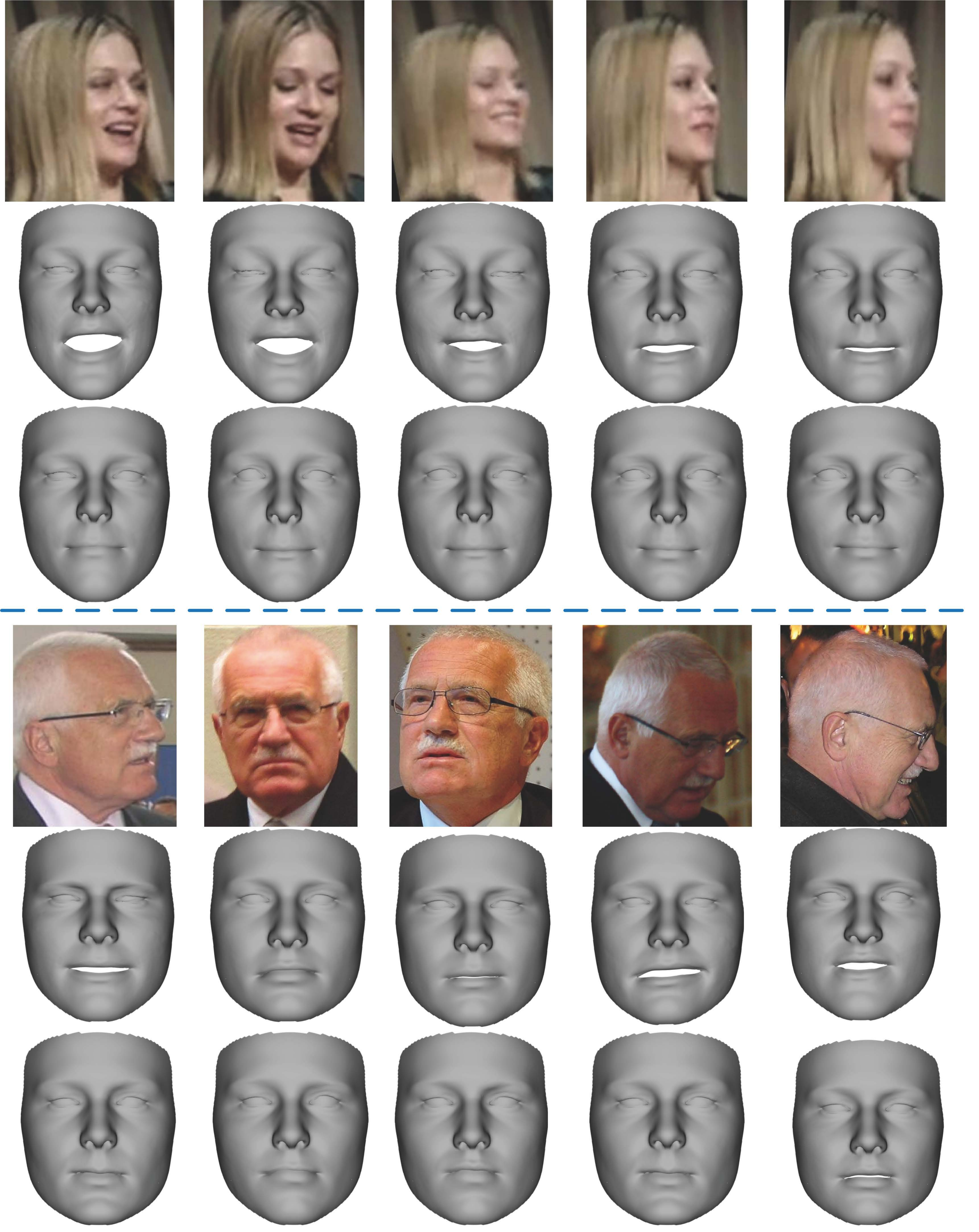}
\end{center}
   \caption{Reconstruction results by our proposed method on images from YTF (top) and IJB-A (bottom). The first row shows the input images, and the second and third rows show the reconstructed 3D shapes and \emph{identity} shapes.}
   \label{fig:performance}
\end{figure*}


\subsection*{Qualitative Results}

\begin{figure}[H]
\begin{center}
   \includegraphics[width=0.91\linewidth]{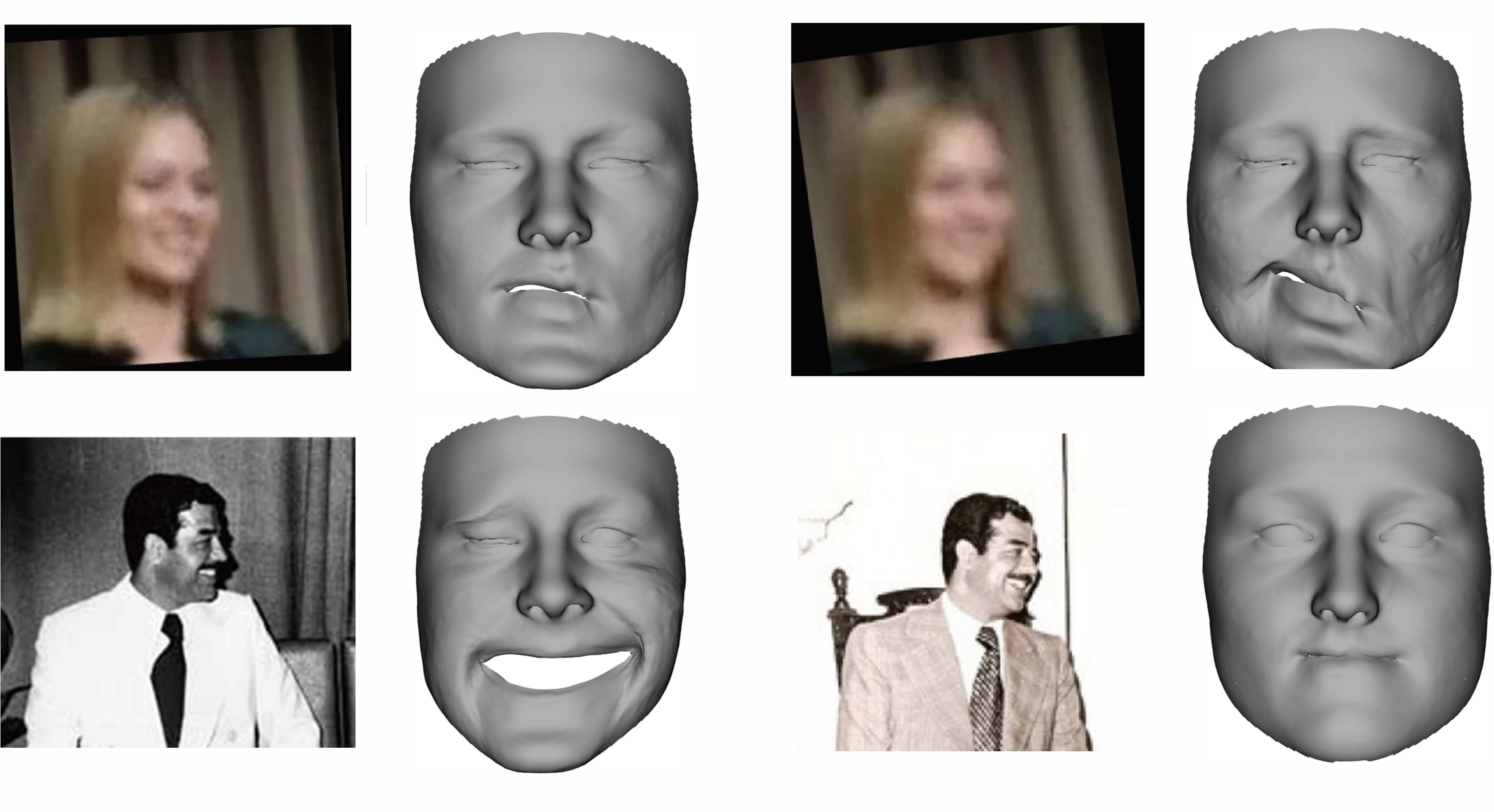}
\end{center}
   \caption{Failure cases of our proposed method due to blurry and very low resolution faces in the images/videos.}
   \label{fig:failure}
\end{figure}

The 3D face reconstruction results of our proposed method on some images from the YTF and IJB-A databases are shown in Figure~\ref{fig:performance}. One can obviously observe from these results that the reconstructed 3D faces do reveal the facial shape deformation (e.g., around the mouth), while the identity shapes successfully disentangle identity-sensitive from identity-irrelevant features. Figure~\ref{fig:failure} shows some images (video frames) for which our proposed method fails to generate plausible 3D face shapes. The blurry and very low resolution faces in these images/videos are the main reasons for the failure.

{\small
\bibliographystyle{ieee}
\bibliography{egbib}

\begin{thebibliography}{10}\itemsep=-1pt

\bibitem{apperID}
\url{https://support.apple.com/en-us/HT208109}.
\newblock Accessed: 2017-11-15.

\bibitem{bagdanov2011florence}
A.~D. Bagdanov, A.~Del~Bimbo, and I.~Masi.
\newblock The florence 2{D}/3{D} hybrid face dataset.
\newblock In {\em Workshop on Human gesture and behavior understanding}, pages
  79--80. ACM, 2011.

\bibitem{bas2016fitting}
A.~Bas, W.~A. Smith, T.~Bolkart, and S.~Wuhrer.
\newblock Fitting a 3{D} morphable model to edges: A comparison between hard
  and soft correspondences.
\newblock In {\em ACCV}, pages 377--391, 2016.

\bibitem{blanz1999morphable}
V.~Blanz and T.~Vetter.
\newblock A morphable model for the synthesis of 3\protect{D} faces.
\newblock In {\em SIGGRAPH}, pages 187--194, 1999.

\bibitem{blanz2003face}
V.~Blanz and T.~Vetter.
\newblock Face recognition based on fitting a 3\protect{D} morphable model.
\newblock {\em TPAMI}, 25(9):1063--1074, 2003.

\bibitem{bowyer2006survey}
K.~W. Bowyer, K.~Chang, and P.~Flynn.
\newblock A survey of approaches and challenges in 3{D} and multi-modal 3{D}+
  {2D} face recognition.
\newblock {\em CVIU}, 101(1):1--15, 2006.

\bibitem{bulat2017far}
A.~Bulat and G.~Tzimiropoulos.
\newblock How far are we from solving the 2{D} \& 3{D} face alignment problem?
  (and a dataset of 230,000 3{D} facial landmarks).
\newblock In {\em ICCV}, 2017.

\bibitem{cao2014facewarehouse}
C.~Cao, Y.~Weng, S.~Zhou, Y.~Tong, and K.~Zhou.
\newblock Facewarehouse: A 3\protect{D} facial expression database for visual
  computing.
\newblock {\em TVCG}, 20(3):413--425, 2014.

\bibitem{cao2016real}
C.~Cao, H.~Wu, Y.~Weng, T.~Shao, and K.~Zhou.
\newblock Real-time facial animation with image-based dynamic avatars.
\newblock {\em TOG}, 35(4):126:1--126:12, 2016.

\bibitem{Dou2017End}
P.~Dou, S.~K. Shah, and I.~A. Kakadiaris.
\newblock End-to-end 3{D} face reconstruction with deep neural networks.
\newblock In {\em CVPR}, 2017.

\bibitem{Emambakhsh2016Nasal}
M.~Emambakhsh and A.~Evans.
\newblock Nasal patches and curves for expression-robust 3{D} face recognition.
\newblock {\em TPAMI}, 39(5):995--1007, 2016.

\bibitem{han20123d}
H.~Han and A.~K. Jain.
\newblock 3\protect{D} face texture modeling from uncalibrated frontal and
  profile images.
\newblock In {\em BTAS}, pages 223--230, 2012.

\bibitem{HanGY17}
X.~Han, C.~Gao, and Y.~Yu.
\newblock Deepsketch2face: A deep learning based sketching system for 3{D} face
  and caricature modeling.
\newblock {\em TOG}, 36(4), 2017.

\bibitem{horn1989shape}
B.~K. Horn and M.~J. Brooks.
\newblock {\em Shape from shading}.
\newblock Cambridge, MA: MIT press, 1989.

\bibitem{huang2007labeled}
G.~B. Huang, M.~Ramesh, T.~Berg, and E.~Learned-Miller.
\newblock Labeled faces in the wild: A database for studying face recognition
  in unconstrained environments.
\newblock Technical report, Technical Report 07-49, University of
  Massachusetts, Amherst, 2007.

\bibitem{jackson2017vrn}
A.~S. Jackson, A.~Bulat, V.~Argyriou, and G.~Tzimiropoulos.
\newblock Large pose 3{D} face reconstruction from a single image via direct
  volumetric {CNN} regression.
\newblock In {\em ICCV}, 2017.

\bibitem{jourabloo2015pose}
A.~Jourabloo and X.~Liu.
\newblock Pose-invariant 3\protect{D} face alignment.
\newblock In {\em ICCV}, pages 3694--3702, 2015.

\bibitem{jourabloo2017ijcv}
A.~Jourabloo and X.~Liu.
\newblock Pose-invariant face alignment via \protect{CNN}-based dense
  3\protect{D} model fitting.
\newblock {\em IJCV}, in press, 2017.

\bibitem{Amin_ICCV17}
A.~Jourabloo, M.~Ye, X.~Liu, and L.~Ren.
\newblock Pose-invariant face alignment with a single cnn.
\newblock In {\em ICCV}, 2017.

\bibitem{kemelmacher20113d}
I.~Kemelmacher-Shlizerman and R.~Basri.
\newblock 3\protect{D} face reconstruction from a single image using a single
  reference face shape.
\newblock {\em TPAMI}, 33(2):394--405, 2011.

\bibitem{kittler1998combining}
J.~Kittler, M.~Hatef, R.~P. Duin, and J.~Matas.
\newblock On combining classifiers.
\newblock {\em TPAMI}, 20(3):226--239, 1998.

\bibitem{Klare2015Pushing}
B.~F. Klare, A.~K. Jain, B.~Klein, E.~Taborsky, A.~Blanton, J.~Cheney,
  K.~Allen, P.~Grother, A.~Mah, and M.~Burge.
\newblock Pushing the frontiers of unconstrained face detection and
  recognition: {IARPA} janus benchmark {A}.
\newblock In {\em CVPR}, pages 1931--1939, 2015.

\bibitem{liu2015cascaded}
F.~Liu, D.~Zeng, J.~Li, and Q.~Zhao.
\newblock Cascaded regressor based 3\protect{D} face reconstruction from a
  single arbitrary view image.
\newblock {\em arXiv:1509.06161}, 2015.

\bibitem{liu2016joint}
F.~Liu, D.~Zeng, Q.~Zhao, and X.~Liu.
\newblock Joint face alignment and 3\protect{D} face reconstruction.
\newblock In {\em ECCV}, pages 545--560, 2016.

\bibitem{Liu2017SphereFace}
W.~Liu, Y.~Wen, Z.~Yu, M.~Li, B.~Raj, and L.~Song.
\newblock Sphereface: Deep hypersphere embedding for face recognition.
\newblock In {\em CVPR}, 2017.

\bibitem{paysan20093D}
P.~Paysan, R.~Knothe, B.~Amberg, S.~Romdhani, and T.~Vetter.
\newblock A 3\protect{D} face model for pose and illumination invariant face
  recognition.
\newblock In {\em AVSS}, pages 296--301, 2009.

\bibitem{peng2017recons}
X.~Peng, X.~Yu, K.~Sohn, D.~N. Metaxas, and M.~Chandraker.
\newblock Reconstruction-based disentanglement for pose-invariant face
  recognition.
\newblock In {\em ICCV}, 2017.

\bibitem{Richardson2016Learning}
E.~Richardson, M.~Sela, R.~Or-El, and R.~Kimmel.
\newblock Learning detailed face reconstruction from a single image.
\newblock In {\em CVPR}, 2017.

\bibitem{romdhani2005estimating}
S.~Romdhani and T.~Vetter.
\newblock Estimating 3\protect{D} shape and texture using pixel intensity,
  edges, specular highlights, texture constraints and a prior.
\newblock In {\em CVPR}, pages 986--993, 2005.

\bibitem{Roth_2016_CVPR}
J.~Roth, Y.~Tong, and X.~Liu.
\newblock Adaptive 3\protect{D} face reconstruction from unconstrained photo
  collections.
\newblock In {\em CVPR}, pages 4197--4206, 2016.

\bibitem{Sela2017Unrestricted}
M.~Sela, E.~Richardson, and R.~Kimmel.
\newblock Unrestricted facial geometry reconstruction using image-to-image
  translation.
\newblock In {\em ICCV}, 2017.

\bibitem{taigman2014deepface}
Y.~Taigman, M.~Yang, M.~Ranzato, and L.~Wolf.
\newblock Deepface: Closing the gap to human-level performance in face
  verification.
\newblock In {\em CVPR}, pages 1701--1708, 2014.

\bibitem{tewari2017mofa}
A.~Tewari, M.~Zollh{\"o}fer, H.~Kim, P.~Garrido, F.~Bernard, P.~P{\'e}rez, and
  C.~Theobalt.
\newblock Mofa: Model-based deep convolutional face autoencoder for
  unsupervised monocular reconstruction.
\newblock In {\em CVPR}, 2017.

\bibitem{tran2016regressing}
A.~T. Tran, T.~Hassner, I.~Masi, and G.~Medioni.
\newblock Regressing robust and discriminative 3{D} morphable models with a
  very deep neural network.
\newblock In {\em CVPR}, 2017.

\bibitem{tran2017disentangled}
L.~Tran, X.~Yin, and X.~Liu.
\newblock Disentangled representation learning gan for pose-invariant face
  recognition.
\newblock In {\em CVPR}, pages 1283--1292, 2017.

\bibitem{LuanPose2017}
L.~Tran, X.~Yin, and X.~Liu.
\newblock Disentangled representation learning \protect{GAN} for pose-invariant
  face recognition.
\newblock In {\em CVPR, in press}, 2017.

\bibitem{Wolf2011Face}
L.~Wolf, T.~Hassner, and I.~Maoz.
\newblock Face recognition in unconstrained videos with matched background
  similarity.
\newblock In {\em CVPR}, pages 529--534, 2011.

\bibitem{yi2013towards}
D.~Yi, Z.~Lei, and S.~Z. Li.
\newblock Towards pose robust face recognition.
\newblock In {\em CVPR}, pages 3539--3545, 2013.

\bibitem{Yi2014Learning}
D.~Yi, Z.~Lei, S.~Liao, and S.~Z. Li.
\newblock Learning face representation from scratch.
\newblock {\em arXiv:1411.7923}, 2014.

\bibitem{yin20063D}
L.~Yin, X.~Wei, Y.~Sun, J.~Wang, and M.~J. Rosato.
\newblock A 3\protect{D} facial expression database for facial behavior
  research.
\newblock In {\em FG}, pages 211--216, 2006.

\bibitem{Yin2017Towards}
X.~Yin, X.~Yu, K.~Sohn, X.~Liu, and M.~Chandraker.
\newblock Towards large-pose face frontalization in the wild.
\newblock In {\em ICCV}, 2017.

\bibitem{Zhang2016Joint}
K.~Zhang, Z.~Zhang, Z.~Li, and Y.~Qiao.
\newblock Joint face detection and alignment using multitask cascaded
  convolutional networks.
\newblock {\em SPL}, 23(10):1499--1503, 2016.

\bibitem{zhu2016CVPR}
X.~Zhu, Z.~Lei, X.~Liu, H.~Shi, and S.~Li.
\newblock Face alignment across large poses: A 3\protect{D} solution.
\newblock In {\em CVPR}, pages 146--155, 2016.

\bibitem{zhu2015high}
X.~Zhu, Z.~Lei, J.~Yan, D.~Yi, and S.~Z. Li.
\newblock High-fidelity pose and expression normalization for face recognition
  in the wild.
\newblock In {\em CVPR}, pages 787--796, 2015.

\end{thebibliography}
}

\end{document}